\documentclass [twocolumn]{svjour3}[]

\usepackage{graphicx}
\usepackage{caption}
\usepackage{verbatim}

\begin{document}

\title{Neuromorphic Robot Dream}
\titlerunning{Neuromorphic Dream}

\author{
Alexander Tchitchigin
\and  Max Talanov
\and Larisa Safina
\and Manuel Mazzara
}

\institute{A. Tchitchigin
  \at Service Science and Engineering Laboratory, Innopolis University
  and Laboratory of Neurobiology, Kazan Federal University, Russia,
\email{a.chichigin@innopolis.ru}
\and M. Talanov
\at Laboratory of Neurobiology, Kazan Federal University, Russia,
\email{max.talanov@gmail.com}
\and L. Safina
\at Service Science and Engineering Laboratory, Innopolis University, Russia,
\email{l.safina@innopolis.ru}
\and M. Mazzara
\at Service Science and Engineering Laboratory, Innopolis University, Russia,
\email{m.mazzara@innopolis.ru}
}

\maketitle

\begin{abstract}

In this paper we present the next step in our approach to neurobiologically plausible
implementation of emotional reactions and behaviors for real-time autonomous robotic systems.
The working metaphor we use is the ``day and the ``night'' phases of mammalian life.
During the ``day phase'' a robotic system stores the inbound information and is controlled by
a light-weight rule-based system in real time. In contrast to that, during the ``night phase''
information that has been stored is transferred to a supercomputing system to update the realistic
neural network: emotional and behavioral strategies.

\keywords{robotics, spiking neural networks, artificial emotions, affective computing}

\end{abstract}

\section{Introduction}\label{intro}

From our perspective emotional mechanisms tend to be as valuable for computational systems
as for human beings. Without understanding and simulation of emotions
the AI--human communication becomes ineffective due to a robotic system making
wrong decisions and inhumane reactions. However, these phenomena seem to be ``difficult''
for computational as well as AI and robotics researchers and require to use the simulation
of nerobiological processes for implementation, expanding the practical autonomous
real-time control of a robotic platform with realistic emotional appraisal and behavior,
based on simulation of spiking neural network (sNN) with neuromodulation.

\section{Day phase and Night phase}\label{my-idea}

\begin{figure}[ht]
  \centering
  \includegraphics[width=0.5\textwidth]{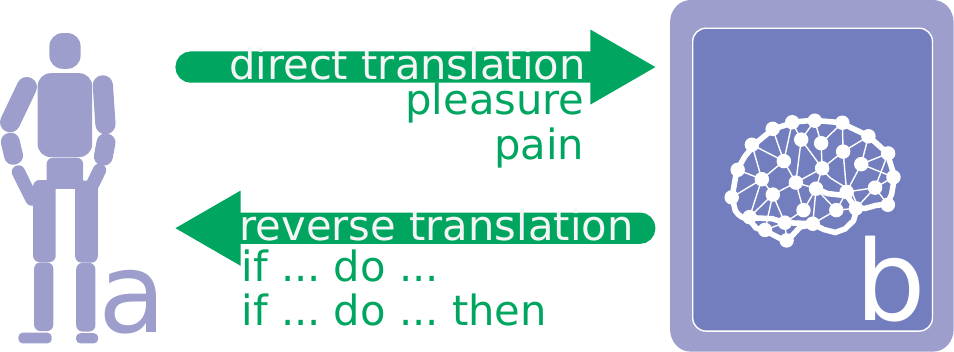}
  \captionsetup{singlelinecheck=off}
  \caption[Neuro Robot Dream]{Neuromorphic robot dream direct and reverse translations}
  \label{fig:neuro-dream}
\end{figure}

We expect real-time behaviour from a robotic system with traditional rule-based control system.
In the ``day phase'' the control system should react on the input stimuli as well as
record and store them for post-processing during the ``night phase''.
We call the supercomputer sNN simulation that does the processing during the ``night phase''
the ``sleeping brain''.
When robot transfers its experience via direct translation,
the ``sleeping brain'' processes its inputs, inferring an emotional
response of the system from levels of neuromodulators~\cite{Vallverdu2015}.
Emotional responses of the sNN are used during the
generation of updates to the rules of the control system which are
sent back to the robot.

\section{Bisimulation and Neuromorphic simulation}\label{the-details}

In our previous paper~\cite{RobotDream2016} we have proposed the ``bisimulation''
mechanism, when we are training the system to adapt to the rules and the structure
of the rules of the particular robotic control system.

The other way is to use synchronisation methods introduced on the figure \ref{fig:neuro-dream}.

%\subsection{Reverse translation}\label{reverse-translation}

When the ``sleeping brain'' is ready to transfer the updated behavioral strategies back to
the robotic system, it starts the procedure that we call ``reverse translation''.
It is based on the mapping of neuronal structures to the rule based description.
This mapping was inherited from the Marvin Minsky's book ``The emotion machine''~\cite{minsky2007}.

Minsky considers rules of the $IF \ldots\ DO \ldots\ THEN \ldots$ shape
that we associate with cortical columns employing models of the sparse distributed
memory~\cite{sparse_memory} and the hierarchical temporal memory~\cite{on_intelligence}.

This temporally consecutive activations correspond to the $IF \ldots\ DO \ldots\ THEN \ldots$
rules where the $THEN$ part represents the expected outcome which is at the same time the condition for
the next rule (following the model by Minsky in~\cite{minsky2007}).

With this approach we could reuse the same ``sleeping brain''
simulation with the number of simultaneously operating robots. The advantages would be
the resource-efficiency (the sNN simulation of a mammalian brain is extremely
memory and CPU consuming) and that the ``sleeping brain'' will
receive more information, more examples and will learn faster.
Also  different robotic systems, possibly
performing different tasks in different environments, will benefit from the each
other experience via the single-brain configuration.

\subsection{Direct translation}\label{sec:direct-translation}

Considering the initial training of the ``sleeping brain'' simulation (nurturing period so to say),
we propose to decouple from a particular control system .
The approach is to perform a translation of raw recorded signals into the
appropriate stimulation of brain cells.
For instance, a robotic system has an optical camera and records an inbound video
stream, the system transforms the stream non-linearly into
the stimulation of visual cortex cells during the direct translation, and different
sensors and channels will use different mapping functions.
The ``pain'' reaction is activated if the signal on any sensory channel is too strong,
triggering the massive neuronal excitation. The proposed mechanism extracts
punishment signals flexibly from raw inbound streams of any robotic system.

We have to admit that the mapping of the ``pleasure'' signals is less clear.
One option is to associate the battery level of a robotic system to the sense
of the ``hunger'' this way the ``charging'' is associated to the feeding thus
``pleasure'' increasing the dopamine level.
Obviously, further research in this direction is needed.

\section{Conclusion}\label{conclusion}

As the development of our previous ideas~\cite{RobotDream2016}
we introduce the integration of a robotic system
with the simulated ``brain'' (a spiking neural network).
We propose the approach of two translations: $robot \rightarrow sNN$ and
$sNN \rightarrow robot$. The first one (``direct translation'') we associate with
the animal training and the second one we call ``reverse translation'', that do not
have direct analogy in biological life. Principles of the direct translation
are the following: we simulate the sensory input triggering the simulated neurones
of proper parts of the brain based on the robotic system ``experience'' that has
been stored, taking into account ``pleasure and pain'' effects. The reverse translation
is based on the multi-pass transformation and convolution of the temporal-probabilistic
rules describing the sNN.

\section{Acknowledgments}
%\begin{acknowledgement}
Part of the work was performed according to the Russian Government Program of Competitive Growth of Kazan Federal University.
%\end{acknowledgement}

%\bibliographystyle{apalike}
\bibliographystyle{spmpsci}
\bibliography{bib2016}

\end{document}